\pdfoutput=1

\documentclass[11pt]{article}
\usepackage{authblk} 

\usepackage{ACL2023}

\usepackage{times}
\usepackage{latexsym}
\usepackage{amsmath}
\usepackage{algorithm}
\usepackage{algpseudocode}
\usepackage{graphicx}
\usepackage{hyperref}
\usepackage{todonotes}
\usepackage{amssymb}
\usepackage{caption}
\usepackage{subcaption}

\usepackage[T1]{fontenc}

\usepackage[utf8]{inputenc}

\usepackage{microtype}

\usepackage{inconsolata}

\title{Finding the Needle in a Haystack:\\ Unsupervised Rationale Extraction from Long Text Classifiers}

\author[1,2]{Kamil Bujel}
\author[2,3]{Andrew Caines}
\author[4,3]{Helen Yannakoudakis}
\author[1,3]{Marek Rei}
\affil[1]{Department of Computing, Imperial College London, U.K.}
\affil[2]{Department of Computer Science and Technology, University of Cambridge, U.K.}
\affil[3]{The ALTA Institute, University of Cambridge, U.K.}
\affil[4]{Department of Informatics, King's College London, U.K.}

\begin{document}
\maketitle
\begin{abstract}
Long-sequence transformers are designed to improve the representation of longer texts by language models and their performance on downstream document-level tasks. However, not much is understood about the quality of token-level predictions in long-form models. We investigate the performance of such architectures in the context of document classification with unsupervised rationale extraction. We find standard soft attention methods to perform significantly worse when combined with the Longformer language model. We propose a compositional soft attention architecture that applies RoBERTa sentence-wise to extract plausible rationales at the token-level. We find this method to significantly outperform Longformer-driven baselines on sentiment classification datasets, while also exhibiting significantly lower runtimes.
\end{abstract}

\section{Introduction}
Transformer-based architectures \citep{vaswani2017attention} have become ubiquitous in natural language processing research. A key attribute to their success is the multi-head self-attention mechanism \citep{michel2019sixteen}. However, its computational and memory requirements grow quadratically with input sequence length. Therefore, models such as BERT \citep{devlin2018bert} commonly limit the maximum sequence length to $512$ tokens. Longer documents are truncated \cite{devlin2018bert} or staggered position embeddings are used \cite{jain-etal-2020-learning}. This limitation motivated long-text transfomers, such as Big Bird \citep{bigbird} and Longformer \citep{longformer}, which reduce the complexity of self-attention through the use of sparse attention and improve the performance of transformers on long documents.

The task of rationale extraction focuses on selecting a subset of input as a justification for the model's output. In particular, our work focuses on extracting token-level rationales from long documents. The models presented are supervised using document labels only, with no token-level annotations used during training. This setting is more practical for longer texts, where token-level annotations are often missing due to prohibitive costs of manual labeling. 

Rationale extraction for the vanilla transformer-based sentence and document classifiers has already been examined \citep{pruthi2020weakly, jain-etal-2020-learning, bujel-etal-2021-zero}. However, to the best of our knowledge, there is no work that evaluates unsupervised rationales extracted from long-text transformers. As we show in this work, methods that work best for standard transformers do not necessarily perform as well on longer documents. We investigate various methods to adapt long-text transformers to zero-shot rationale extraction. We highlight that this work explicitly focuses on extracting \textit{plausible} rationales (agreeable to human annotators; \citet{deyoung-etal-2020-eraser}), as opposed to \textit{faithful} explanations (true to the system's computation; \citet{rudin2018please}).

While self-attention layers have been shown to learn correct grammatical relations \citep{clark2019does} and rationales \citep{pruthi2020weakly, jain-etal-2020-learning}, we find that the sparse self-attention present in Longformer struggles to locate tokens that can serve as a plausible rationale for document labels. For longer sentiment detection datasets, the quality of the rationale extracted from the self-attention layers is not better than the random baseline.  We further show that a direct combination of Longformer with a Weighted Soft Attention rationale extractor \cite{bujel-etal-2021-zero} does not perform well either, as only a small proportion of tokens receive supervision signal at each epoch. We propose Ranked Soft Attention, which ensures that the weights of all tokens are updated at each epoch. We introduce a Compositional Soft Attention architecture, which applies a vanilla RoBERTa model sentence-wise \citep{liu2019roberta} and composes the obtained token contextual embeddings using Ranked Soft Attention. We find this model to outperform other unsupervised approaches for rationale extraction on the longer sentiment detection documents, while being significantly faster than Longformer-based systems.

\section{Soft Attention}

\citet{rei2018zero} introduced a soft attention architecture for biLSTM zero-shot sequence labelers, which \citet{bujel-etal-2021-zero} adapted to transformers by introducing Weighted Soft Attention. We apply Weighted Soft Attention to contextual embeddings $t_i \in \mathbb{R}^{h}$ for all tokens in the document: 
\begin{align}
    \label{eq:soft-attn}
    e_i = \tanh(W_{e}t_i + b_{e})\,\,\,\,\,\,\,\,\, \\
    \widetilde{a_i} = \sigma(\widetilde{e_i}) \,\,\,\,\,\,\,\,\,\,\widetilde{e_i} = W_{\tilde{e}}e_i + b_{\tilde{e}}
\end{align}
where $h \in \mathbb{N}$ is the dimension of the contextual token embeddings, $e_i \in \mathbb{R}^{h'}$ is a vector, $\widetilde{e_i} \in \mathbb{R}$ is a single scalar value and $\widetilde{a_i} \in [0, 1]$ is the token attention score. The scores are converted to normalised attention weights $a_i$ to build document-level representation $c$:
\begin{equation}
    a_i = \frac{\widetilde{a_i}^\beta}{\sum_{j=1}^N \widetilde{a_j}^\beta} \,\,\,\,\,\,\,\,\,\,
    c = \sum_{i=1}^N a_it_i
\end{equation}
\begin{equation}
    d = \tanh(W_dc+b_d)
    \,\,\,\,\,\,\,\,\,\,
    y = \sigma(W_yd+b_y)
\end{equation}
where $N \in \mathbb{N}$ is the number of tokens in a sentence, $c \in \mathbb{R}^h$ is the combined document representation, $d \in \mathbb{R}^s$ is the hidden document representation and $y \in \mathbb{R}^2$ is the document prediction. $\beta \in \mathbb{R}$ is a weight controlling the sharpness of the attention scores. For each document $j$, we obtain document-level predictions $y^{(j)} \in [0, 1]$ and the token-level scores $0 \leq \widetilde{a_i} \leq 1$. As the token-level scores are preceded by a logistic activation function, we use a classification threshold of $0.50$. The supervision of Weighted Soft Attention is described in Appendix \ref{apdx:W-SA}.

\subsection{Ranked Soft Attention}
Empirically, we find that most of the token scores obtained using Weighted Soft Attention are close to $1$ for long documents. We suspect this is caused by an insufficient number of token scores receiving a supervision signal at each epoch. Only minimum and maximum token scores are optimized. While working well for individual sentences, such an approach does not scale well to longer documents. We confirmed this hypothesis by measuring that on average, only 5\% of tokens receive supervision signal during training.

    We propose Ranked Soft Attention, where $k\%$ of tokens with the highest scores are supervised with the document label, while the remaining $100-k\%$ tokens are supervised with a $0$. In particular, we define two subsets, $\widetilde{a}_{top}$ and $\widetilde{a}_{rest}$ which contain the top $k\%$ and the remaining $1 - k\%$ of token attention scores $\widetilde{a_i}$ respectively, so that $\widetilde{a} = \widetilde{a}_{top} + \widetilde{a}_{rest}$. We define the loss function $L_{ranked}$ as follows:

\begin{multline}
    L_{ranked} = \sum_j \Bigg(\frac{1}{|\widetilde{a}_{top}|}\sum_{\widetilde{a_i} \in \widetilde{a}_{top}} (\widetilde{a_i} - \widetilde{y}^{(j)})^2 \\\\
    + \frac{1}{|\widetilde{a}_{rest}|}\sum_{\widetilde{a_i}\in \widetilde{a}_{rest}} (\widetilde{a_i} - 0)^2 
    \Bigg)
\end{multline}
where $k$ is a hyperparameter that can be inferred based on the percentage of annotations in the dataset \citep{jain-etal-2020-learning} and total loss is $L = L_1 + \gamma(L_2 + L_3) + \gamma_{ranked} L_{ranked}$. 

We hope this universal supervision approach alleviates the problems of insufficient token-level supervision during training.

\begin{table*}[t]
\centering
\setlength{\tabcolsep}{2.3pt}
\small
\begin{tabular}{l|cccc|c|cccc|c}
 & \multicolumn{5}{c|}{FCE} & \multicolumn{5}{c}{BEA 2019} \\
 & Doc $F_1$ & $F_1$ & $F_{0.5}$ & MAP & Time & Doc $F_1$ & $F_1$ & $F_{0.5}$ & MAP  & Time \\
 \hline
 Random Uniform  & -  & $12.13 \pm 0.36$ & 13.70 & 15.91  & - & - & $11.36 \pm 0.32$ & 12.00 & 16.68  &  - \\
 \hline
Longformer Weighted Soft Attention & \textbf{89.34} & $\textbf{24.81} \pm 1.42$ & 17.64 & 16.17  & 242 & 93.79 & $\textbf{21.22} \pm 0.17$ & 14.59 & 16.38  & 357 \\
Longformer Self-Attention Top-K & 89.29 & $14.23 \pm 2.25$ & 15.57 & 14.67  & 243 & \textbf{94.39} & $14.75 \pm 3.21$ & 14.99 & 13.76  & 360 \\
\hline
Longformer Ranked Soft Attention & 89.10 & $22.23 \pm 2.44$ & \textbf{23.40} & 18.82  & 238 & 93.49 & $19.11 \pm 2.92$ & 19.65 & 18.55  & 351 \\
Compositional Soft Attention & 81.32 & $21.08 \pm 2.02$ & 23.19 & \textbf{19.67}  & \textbf{139} & 89.07 & $20.44 \pm 1.73$ & \textbf{20.22} & \textbf{19.44}  & \textbf{141} \\
\hline
\end{tabular}
\caption{Results for the Grammatical Error Detection Datasets. Doc $F_1$ represents document-level classification performance, while $F_1$, $F_{0.5}$ and \textit{MAP} represent token-level metrics. \textit{Time} represents an average seconds per epoch the given model took to train. We see that both our Ranked Soft Attention and Compositional Soft Attention perform significantly better on the token-level than Weighted Soft Attention or Self-Attention baselines. We note that Weighted Soft Attention exhibits high $F_1$ due to assigning $1$ to most tokens and thus exhibiting a large recall. Compositional Soft Attention is significantly faster than Longformer-based models.}
\label{tab:ged}
\end{table*}

\subsection{Compositional Soft Attention}
Given RoBERTa's strong performance as a rationale extractor for individual sentences \cite{bujel-etal-2021-zero, pruthi2020weakly}, we decide to investigate the feasibility of using RoBERTa together with Ranked Soft Attention to extract token-level rationales from longer documents. In particular, we propose to model intra-sentence token dependencies by applying RoBERTa to each sentence individually. To extract rationale for the whole document, we use a ranked soft attention layer that composes the individual token contextual embeddings across different sentences. This is different from Hierarchical Transformers \citep{pappagari2019hierarchical}, which focus on document-level representation only.

More concretely, we propose a Compositional Soft Attention architecture that uses a standard length transformer to build contextual token embeddings $t_{k}' \in \mathbb{R}^{N_k\times d}$ separately for each sentence $s_k$, where $k \in \mathbb{N}$ and $k < m$:

\begin{equation}
    t_k' = \text{Transfomer}(s_k)
\end{equation}
\begin{equation}
    t = \text{Concat}[t_0', ..., t_m']
\end{equation}
where $t \in \mathbb{R}^{N \times d}$. We then provide this document representation $t$ as input to a soft attention layer (Eq.\ref{eq:soft-attn}), which composes tokens across all sentences to obtain a document-level representation and the prediction $y^{(j)}$. We additionally obtain token-level attention scores $\widetilde{a_i}$, which we use to extract rationales for document classification tasks. An architecture visualization and pseudocode are given in Appendix \ref{apdx:compositional}.


\section{Datasets}
We investigate the performance of our models on three different datasets. Each of the datasets contains a document-level label, together with human annotations on the token-level that are used for evaluation only.

We evaluate our models on Grammatical Error Detection (GED) datasets, which contain texts written by learners of the English language. They are annotated with token-level grammatical errors, which serve as rationale for document-level proficiency scores. An annotated set of essays from the Write \& Improve feedback and assessment platform \cite{yannakoudakis2018developing} was released as part of the \textbf{BEA 2019}\footnote{\href{https://www.cl.cam.ac.uk/research/nl/bea2019st/}{https://www.cl.cam.ac.uk/research/nl/bea2019st/}} shared task \cite{bryant-etal-2019-bea}. The essays were submitted online in response to various prompts, and document-level labels indicate the CEFR proficiency levels (A/B/C) of the English learners. We remove intermediate (B) essays and treat the beginner (A) class as a positive document label, as opposed to the advanced (C) class label. Since there is no publicly available test dataset, we held out the development dataset for evaluation purposes and randomly sampled $10\%$ of the training dataset for development. 

The First Certificate in English\footnote{\href{https://ilexir.co.uk/datasets/index.html}{https://ilexir.co.uk/datasets/index.html}} \textbf{(FCE)} dataset \cite{yannakoudakis2011new} contains essays written by non-native English learners for an intermediate-level language proficiency exam. Each student wrote $2$ essays, which we combine to increase the overall document length. We split the dataset into beginner (\textit{score}\,\,$<27$, equivalent to a fail) and more advanced learners (\textit{score}\,\,$>30$). We use the train/dev/test split released by \citet{rei2016compositional}. We note that both of the GED datasets contain a relatively small proportion of documents exceeding $512$ tokens, the standard RoBERTa maximum sequence length (Table \ref{tab:data-statistics}).

We also use the Sentiment Detection movie reviews IMDB\footnote{\href{https://www.tensorflow.org/datasets/catalog/movie_rationales}{https://www.tensorflow.org/datasets/catalog/movie\_rationales}} dataset \cite{zaidan2007using}, which contains positive and negative movie reviews. We focus on a subset of this dataset that has been annotated with rationales for the reviews by human annotators.  We split the dataset into \textbf{IMDB-Pos} and \textbf{IMDB-Neg}, where the former contains evidence for only positive reviews, while the latter contains evidence for negative ones. We use the train/dev/test split published by \citet{pruthi2020weakly}.

\begin{table*}[t]
\centering
\setlength{\tabcolsep}{2.3pt}
\small
\begin{tabular}{l|cccc|c|cccc|c}
                                & \multicolumn{5}{c|}{IMDB-Pos}                                                                                          & \multicolumn{5}{c}{IMDB-Neg}                                                                                         \\
                                
                                & \multicolumn{1}{c}{Doc $F_1$} & \multicolumn{1}{c}{$F_1$} & \multicolumn{1}{c}{$F_{0.5}$} & \multicolumn{1}{c}{MAP} & \multicolumn{1}{|c|}{Time} & \multicolumn{1}{c}{Doc $F_1$} & \multicolumn{1}{c}{$F_1$} & \multicolumn{1}{c}{$F_{0.5}$} & \multicolumn{1}{c}{MAP} & \multicolumn{1}{|c}{Time} \\
                                \hline
Random Uniform         & -     & $5.46 \pm 0.18$                    & 4.27                      & 8.46 &          -            & -       & $6.02 \pm 0.23$                     & 4.80                       & 9.90 &  -  \\
\hline
Longformer Weighted Soft Attention & 93.41                    & $6.89 \pm 0.19$                     & 4.43                       & 7.81 & 603                      & 93.52                    & $8.34 \pm 0.98$                    & 5.40                       & 10.59 & 599                     \\
Longformer Self-Attention Top-K  & \textbf{94.42}           & $5.90 \pm 2.21$                     & 5.19                       & 7.85 & 606                      & \textbf{94.60}           & $5.71 \pm 3.63$                     & 5.02                       & 9.03 & 603                     \\
\hline
Longformer Ranked Soft Attention   & 92.63                    & $14.13 \pm 2.27$                   & 11.92                      & 11.97 & 618                     & 93.92                    & $19.62 \pm 0.61$                    & 16.98                      & 16.44  & 608                   \\
Compositional Soft Attention       & 91.85                    & $\textbf{25.46} \pm 0.97$           & \textbf{20.12}            & \textbf{26.82} & \textbf{436}            & 90.82                    & $\textbf{27.27} \pm 3.68$         & \textbf{22.70}             & \textbf{29.78} & \textbf{433}           \\       
\hline
\end{tabular}
\caption{Results for the Sentiment Detection IMDB datasets. Doc $F_1$ represents document-level classification performance, while $F_1$, $F_{0.5}$ and \textit{MAP} represent token-level metrics. \textit{Time} represents an average seconds per epoch the given model took to train. We note the low performance of previously used Weighted Soft Attention and Longformer Self-Attention. Both of our Ranked Soft Attention and Compositional Soft Attention perform significantly better on the rationale extraction. Compositional Soft Attention performs best across all token-level metrics, while also exhibiting lower runtimes.}
\label{tab:imdb}
\end{table*}

\section{Results}
We report the $F_1$-measure and the $F_{0.5}$-measure on the token level. We further evaluate the models in the ranking setting by using Mean Average Precision (MAP) for returning positive tokens. MAP serves as a diagnostic metric that removes the threshold dependency and indicates which models have learned to rank the tokens best. We further report a document-level $F_1$-measure.
 
 As a baseline, we use the global attention scores from the \texttt{<CLS>} token in the Longformer's last self-attention layer. We classify top $k\%$ of token scores as rationale, where the value of $k$ is based on the rationale proportion in the dataset. This is equivalent to evaluating the plausiblity of rationale extracted by the support model in FRESH \citep{jain-etal-2020-learning}.

 Details of our experimental setup are described in Appendix \ref{apdx:experimental-setup}. Tables \ref{tab:ged} and \ref{tab:imdb} present results for the Grammatical Error Detection and Sentiment Detection datasets respectively. 

We find that both the Longformer Self-Attention and Longformer Weighted Soft Attention perform poorly on the task of token-level rationale extraction. On the longer Sentiment Detection dataset, the performance of both methods is on-par with a random baseline. For GED, we note that the high token-level $F_1$ score of Weighted Soft Attention  is due to the model assigning scores of $1$ to most tokens, as evident by the substantially lower $F_{0.5}$ score and high recall (Appendix \ref{apdx:results}). We suspect this homogeneity of token scores is caused by only $2$ tokens per document receiving supervision signal each epoch. This encourages the token scores to stay close to the initial values and the model is unable to learn to provide plausible rationales for its predictions. Increasing the supervision signal through Ranked Soft Attention significantly improves the token-level $F_{0.5}$ results ($5.06\% - 11.58\%$ absolute increase).

We suspect the poor performance of Longformer Self-Attention is partly caused by the use of normalized global attention from the \texttt{<CLS>} token. As the token attention scores across the whole document have to sum up to $1$, very few tokens would be assigned high scores. This is evidenced by the significantly lower token-level recall than for other methods. While some improvement could be obtained through fine-tuning the classification threshold, self-attention methods also do not learn to rank tokens well, as indicated by the low MAP scores, which are below the random baseline.

We notice that Compositional Soft Attention models exhibit significantly lower runtimes ($30\% - 60\%$) than Longformer-based methods. It significantly outperforms all architectures on the longer IMDB datasets, while also achieving results on-par with Ranked Soft Attention on the GED datasets. It appears that the compositional nature of this architecture allowed the RoBERTa model to learn to provide meaningful token-level scores for each sentence individually. This is in contrast with Longformer-based methods, which directly focus on modeling global dependencies and struggle to provide meaningful token-level predictions. We provide example token-level predictions in Appendix \ref{apdx:eg-preds}.

\section{Conclusion}
We investigated unsupervised rationale extraction for long document classifiers. Our experiments showed that standard transformers methods do not perform well on longer texts. We proposed Ranked Soft Attention that works well in conjuction with Longformer by increasing the supervision signal available to each token. We further introduced a novel Compositional Soft Attention architecture that uses RoBERTa to support documents of arbitrary length. We found Compositional Soft Attention to significatly outperform Longformer-based systems on longer documents, while being $30\%-60\%$ faster to fine-tune. 

\clearpage

\section*{Limitations}
Our work aims to bridge the gap between rationale extraction for standard length transformers and long documents. We improve the scalability of such methods by sequentially applying RoBERTa to each sentence instead of relying on slower long text transformers. However, it is important to underline that this method still doesn't not permit application to texts of arbitrary length, as the memory of the GPU is the main limitation. We hope for future work to address this.

We also note the limited evaluation for truly long documents. This is due to small number of long text datasets with token-level annotations available. We encourage the development of such new datasets in the future.

We believe that current approaches of evaluating rationale extraction as a sequence labeling problem do not scale well to longer documents. It is becoming harder to quantitatively evaluate such models. We encourage future work to investigate alternative methods of evaluating plausibility of token-level rationale extractors for long texts.

\clearpage
\bibliography{anthology,custom}
\bibliographystyle{acl_natbib}

\clearpage

\appendix

\section{Supervision of Weighted Soft Attention}
\label{apdx:W-SA}

We further recall that the Weighted Soft Attention architecture uses the following loss functions:

\begin{align}
        L_1 &= \sum_j (y^{(j)} - \tilde{y}^{(j)})^2 \label{l1-redone} \\
    L_2 &= \sum_j (min(\widetilde{a_i}) - 0)^2 \label{l2-redone} \\
    L_3 &= \sum_j (max(\widetilde{a_i}) - \tilde{y}^{(j)})^2 \label{l3-redone}\\
    L &= L_1 + \gamma(L_2 + L_3)\label{loss-zero-shot-redone}
\end{align}
where $L_1$ optimises the document-level performance, $L_2$ ensures the minimum attention score is close to $0$ and $L_3$ optimises the maximum attention score to be close to the document label $\widetilde{y}^{(j)}$.

\section{Compositional Soft Attention}
\begin{figure*}[ht!]
    \centering
    \includegraphics[scale=0.40]{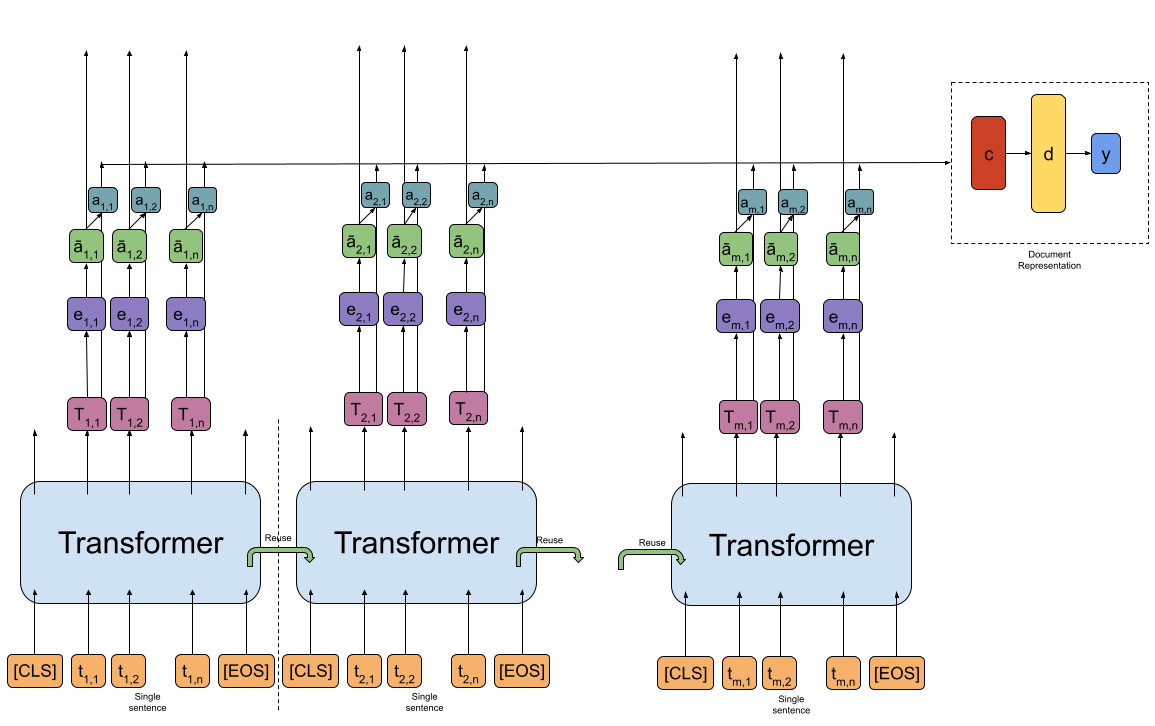}
    \caption{Compositional Soft Attention architecture for rationale extraction and document classification. A standard size transformer is applied to each sentence individually and the contextual token embeddings are then combined to build a document-level representation.}
    \label{fig:composition-visual}
\end{figure*}

\label{apdx:compositional}
We present the algorithm for Compositional Soft Attention in Algorithm \ref{alg:composition} and the overview of the architecture in Figure \ref{fig:composition-visual}.

\begin{algorithm}[h!]
    \footnotesize
    \caption{Compositional Soft Attention}
    \label{alg:composition}
    \begin{algorithmic}
        \For {sentence $s_{j,i}$ in document $doc_j$}
            \State $T_{j,i} \gets \text{Transformer}(s_{j,i})$
        \EndFor
        
        \State $T_j \gets [T_{j,1}, ..., T_{j,n}]$
        \State $\tilde{y}^{(j)},\,\, \widetilde{a} = \text{SoftAttention}(T_j)$
    \end{algorithmic}
\end{algorithm}
\newpage
\section{Datasets}
\label{apdx:datasets}
We present the summary statistics for the datasets used in our experiments in Table \ref{tab:data-statistics}.

\section{Experimental Setup}
\label{apdx:experimental-setup}
We use a pre-trained RoBERTa-base \cite{liu2019roberta} and Longformer-base \cite{longformer} models, available through the HuggingFace library \cite{wolf2019huggingface}. All experiments are performed on Nvidia Tesla P100. Following \citet{mosbach2020stability}, we train for $20$ epochs, with the best performing checkpoint chosen. Each experiment is repeated $3$ times and the average results are reported. We perform significance testing using a two-tailed paired t-test ($a = 0.05$). We set $k$ based on the percentage of evidence present for the positive class.
\subsection{Baselines}
For reference, we include a random baseline, which samples token-level scores from the standard uniform distribution and uses a threshold $k$ to classify the scores. We also compare against a Longformer's last self-attention layer, following insight by \citet{jain-etal-2020-learning} that self-attention performs well as a rationale extractor.

\section{Full Results}
\label{apdx:results}
We present full results of the experiments in Tables \ref{tab:full_bea}, \ref{tab:full_fce}, \ref{tab:full_imdb_pos} and \ref{tab:full_imdb_neg}.

\section{Example Predictions}
\label{apdx:eg-preds}
\begin{figure*}
    \centering
    \begin{subfigure}[t]{0.45\textwidth}
            \centering
        \includegraphics[scale=0.3]{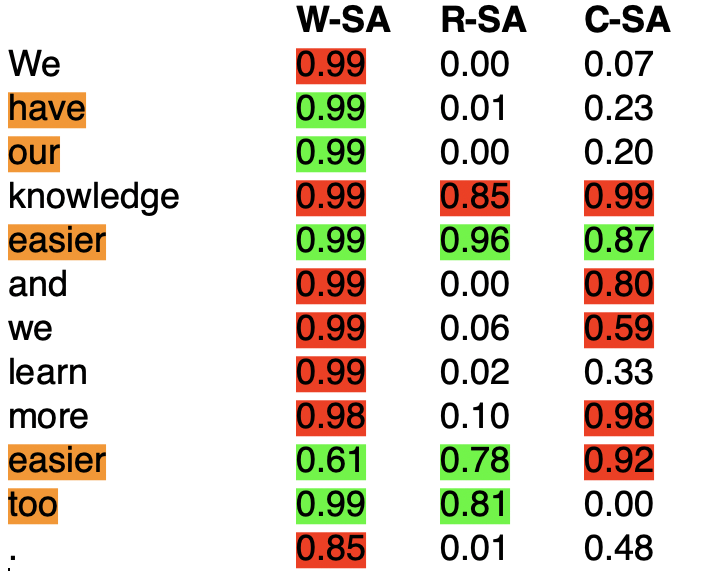}
        \caption{Sample token-level predictions for BEA 2019 positive sample (beginner learner). Compositional Soft Attention finds all evidence, but also scores neighboring tokens highly. Ranked Soft Attention on the other hand attends to less neighbouring tokens. We note that this might explain the performance differences between Grammatical Error Detection and Sentiment Detection datasets, as the the former annotations are more concentrated than the latter.}
        \label{fig:my_label}
    \end{subfigure}
    \hfill
    \begin{subfigure}[t]{0.45\textwidth}
        \centering
        \includegraphics[scale=1.0]{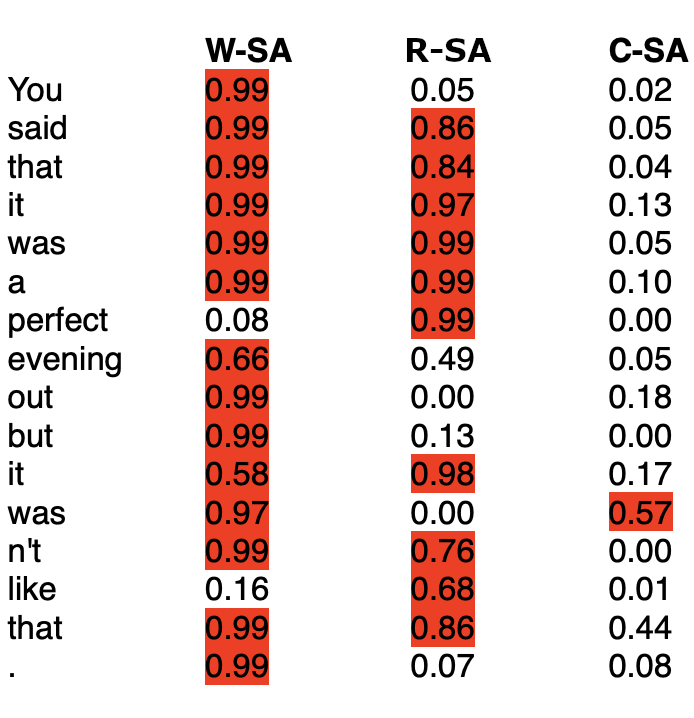}
        \caption{Excerpt from an FCE positive document (beginner learner) without any grammatical errors in the sentence. We note that both Weighted Soft Attention and Ranked Soft Attention find a lot of false positives. On the other hand, Compositional Soft Attention correctly does not much rationale.}
    \end{subfigure}
    
    \begin{subfigure}[b]{0.45\textwidth}
        \centering
        \includegraphics[scale=0.9]{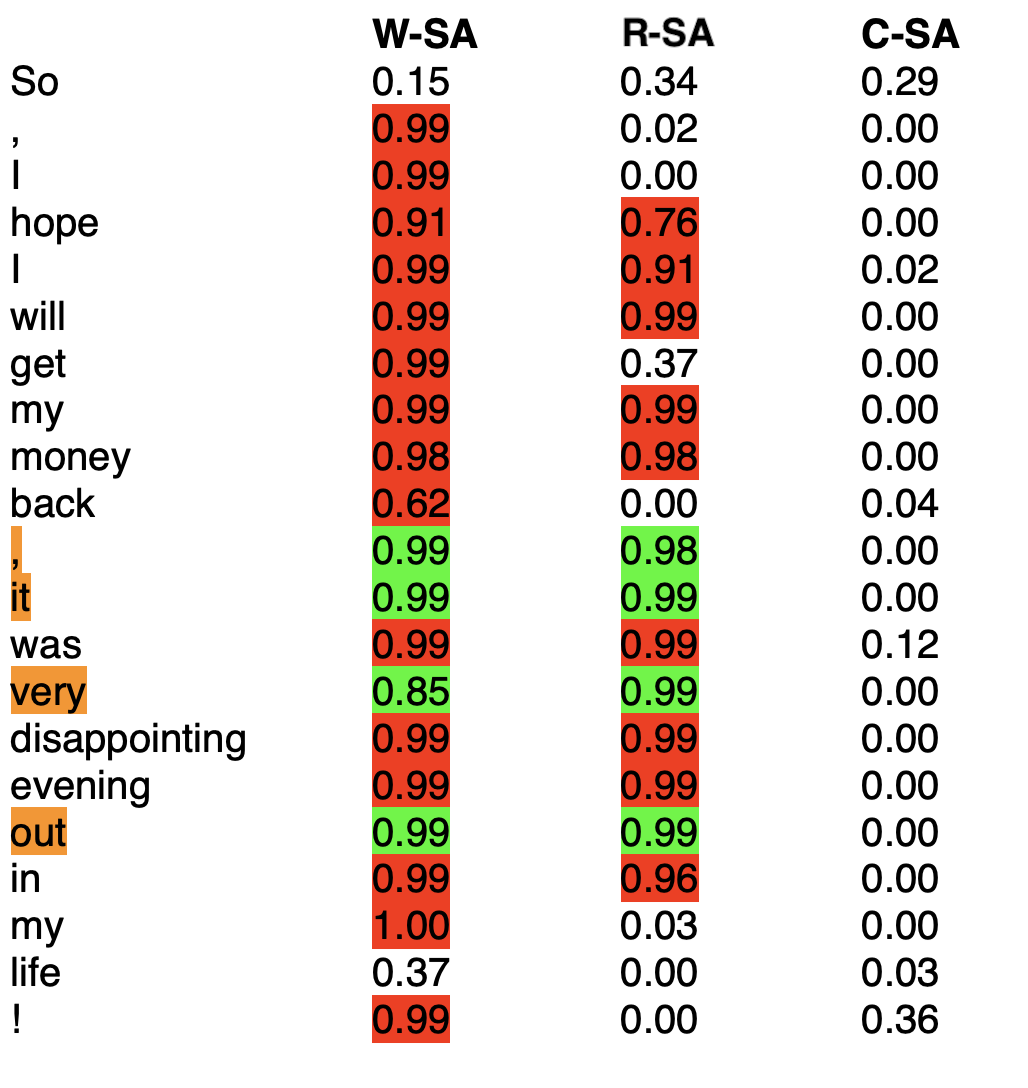}
        \caption{Excerpt from a positive FCE document (beginner learner). This sentence includes grammatical errors. Ranked Soft Attention manages to pick up some manually annotated rationale, while Compositional Soft Attention fails. However, Ranked Soft Attention also finds many false positives. }
    \end{subfigure}
    \hfill
    \begin{subfigure}[b]{0.45\textwidth}
        \centering
        \includegraphics[scale=0.9]{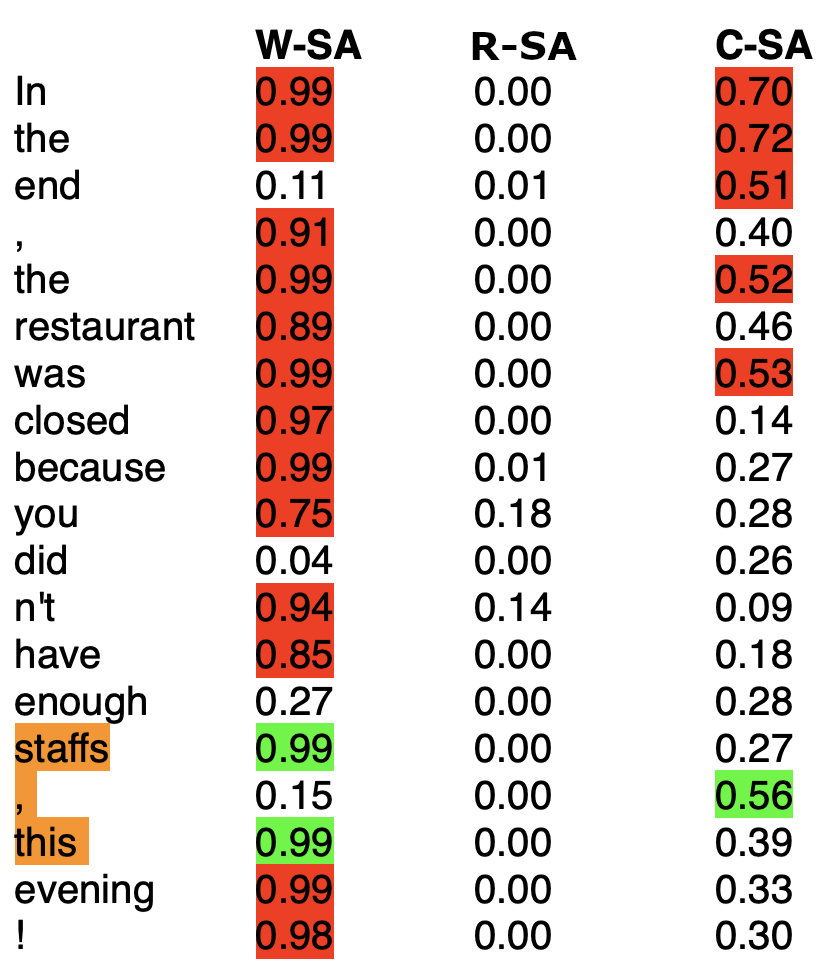}
        \caption{An excerpt from a FCE positive document (beginner learner), for a sentence with grammatical errors. Ranked Soft Attention fails to provide any rationale, while Compositional Soft Attention correctly finds \textit{","} as rationale, but fails to attend to neighboring tokens. }
    \end{subfigure}
    
    \caption{Example predictions for Grammatical Error Detection datasets. \textbf{W-SA} corresponds to Weighted Soft Attention, \textbf{R-SA} to Ranked Soft Attention, while \textbf{C-SA} to Compositional Soft Attention. We highlight words that human annotators marked as \colorbox{orange}{rationale}, while also marking \colorbox{green}{true positives} and \colorbox{red}{false positives}. }
    \label{fig:eg-ged}
\end{figure*}

\begin{figure*}
    \begin{subfigure}[t]{0.45\textwidth}
        \centering
        \includegraphics[scale=0.4]{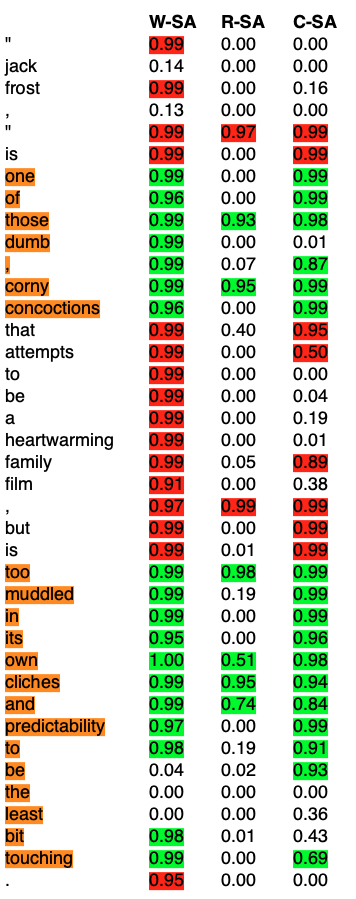}
        \caption{Sample predictions for a sample negative review in the IMDB-Neg dataset. We find that Weighted Soft Attention assigns similar scores to most tokens, while Ranked Soft Attention and Compositional Soft Attention manage to provide more fine-grained predictions. Compositional Soft Attention appears to recover spans of rationale better than Ranked Soft Attention.}
    \end{subfigure}
    \hfill
    \begin{subfigure}[t]{0.45\textwidth}
        \centering
        \includegraphics[scale=0.50]{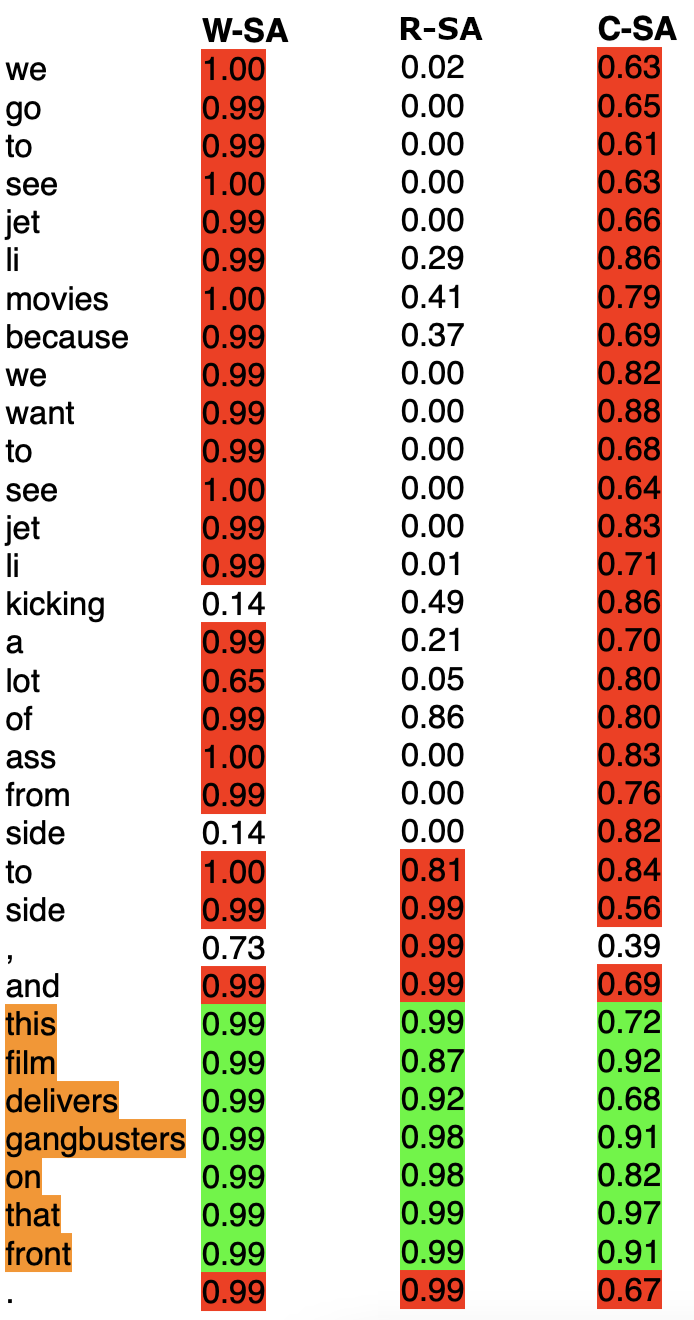}
        \caption{Sample prediction for a positive movie review in the IMDB-Pos review. Here we note how Compositional Soft Attention learns the correct ranking of tokens, as evidenced by the higher scores for true positives than false positives. However, it still failed to optimize correctly for the classification threshold, leading numerous false positives. This problem is not present in the Ranked Soft Attention for this sample.}
    \end{subfigure}

    \centering
    \begin{subfigure}[b]{0.5\textwidth}
        \centering
        \includegraphics[scale=1.0]{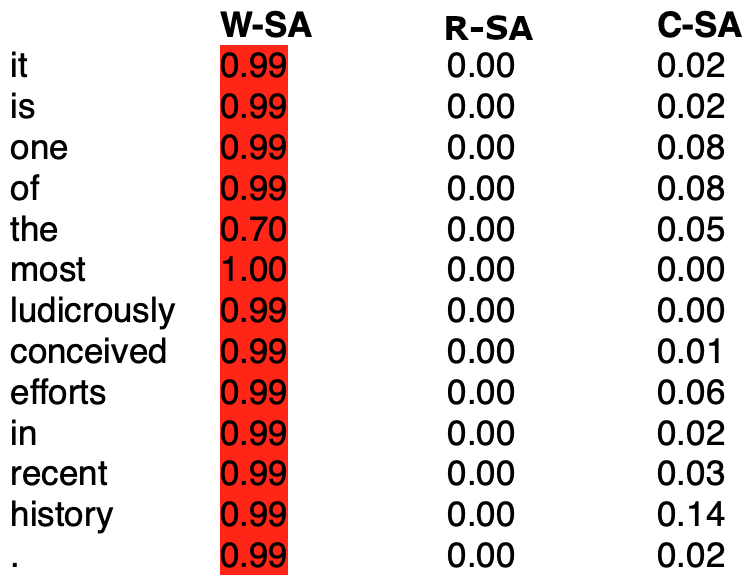}
        \caption{Sample predictions for a negative review in the IMDB-Pos dataset. The gold token-level labels are all $0$, as there are no positive rationale in the negative review. Weighted Soft Attention still assigns scores close to $1$ to most tokens, while both Ranked Soft Attention and Compositional Soft Attention learns not to attend to any tokens. This shows how the increased token-level supervision signal helps these architectures to learn to provide better token-level rationale.}
    \end{subfigure}
    \caption{Example predictions for Sentiment Detection IMDB datasets. \textbf{W-SA} corresponds to Weighted Soft Attention, \textbf{R-SA} to Ranked Soft Attention, while \textbf{C-SA} to Compositional Soft Attention. We highlight words that human annotators marked as \colorbox{orange}{rationale}, while also marking \colorbox{green}{true positives} and \colorbox{red}{false positives}.}
    \label{fig:eg-sentiment}
\end{figure*}

Furthermore, we provide more sample predictions made by different models in Figures \ref{fig:eg-ged} and \ref{fig:eg-sentiment}.

\begin{table*}[h!]
    \centering
    \begin{tabular}{l|cccc}
         & FCE & BEA 2019 & IMDB-Pos & IMDB-Neg  \\
         \hline
         Number of train samples & $722$ & $1120$ & $1200$ & $1200$ \\
         Number of dev samples & $51$ & $280$ & $299$ & $299$ \\
         Number of test samples & $66$  & $200$ & $300$ & $300$  \\
         \hline
         Average text length (words) & $441$ & $213$ & $686$ & $686$ \\
         Maximum text length (words) & $725$ & $655$ & $1935$ & $1935$ \\
         \% of texts $>512$ words & $16\%$ & $2\%$ & $73\%$ & $73\%$ \\ 
          \% positive samples & $49\%$ & $46\%$ & $50\%$ & $50\%$ \\
         \% negative samples & $51\%$ & $54\%$ & $50\%$ & $50\%$\\
         \hline
         \% evidence & $13\%$ & $9\%$ & $8\%$ & $8\%$ \\
         \hline
    \end{tabular}
    \caption{Statistics for the datasets used. All measured on the development datasets. We note the low proportion of long texts in FCE and BEA 2019.}
    \label{tab:data-statistics}
\end{table*}

\clearpage

\begin{table*}[h!]
    \centering
    \begin{tabular}{l|cccccc}
         & Doc $F_1$ & $F_1$ & $F_{0.5}$ & $P$ & $R$ & MAP \\
         \hline 
        Random Uniform & - & $11.36 \pm 0.32$  & $12.00$   & $12.47$	 & $10.42$ & $16.68$ \\
        \hline
        Longformer Weighted Soft Attention & $93.79$    & $\textbf{21.22} \pm 0.17$ & $14.59$ & $12.08$	 &$ \textbf{88.38}$ & $16.38$ \\
        Longformer Soft Attention Top-K & $\textbf{94.39}$    & $14.75 \pm 3.21$ & $14.99$ & $15.16$	 & $14.40$  & $13.76$ \\
        \hline
        Longformer Ranked Soft Attention & $93.49$    & $19.11 \pm 2.92$ & $19.65$ & $\textbf{20.12}$	 &  $18.53$ & $18.55$ \\
        Compositional Soft Attention & $89.07$    & $20.44 \pm 1.73$ & $\textbf{20.22}$ & $20.10$	 & $20.95$ & $\textbf{19.44}$ \\
        \hline
    \end{tabular}
    \caption{Full results for BEA 2019. We note that while Weighted Soft Attention performs best on the token-level, it is largely due to the model assigning scores of $1$ to most tokens, as indicated by the high recall. Using $F_{0.5}$ as an evaluation metric highlights this issue. Our proposed Compositional Soft Attention performs best on the token-level in terms of both $F_{0.5}$ and MAP.}
    \label{tab:full_bea}
\end{table*}

\begin{table*}[]
    \centering
    \begin{tabular}{l|cccccc}
         & Doc $F_1$ & $F_1$ & $F_{0.5}$ & $P$ & $R$ & MAP \\
         \hline 
        Random Uniform & - & $12.13 \pm 0.36$   & $13.70$ & $14.99$	 & $10.18$ & $15.91$ \\
        \hline
        Longformer Weighted Soft Attention & $\textbf{89.34}$    & $\textbf{24.81} \pm 1.42$ & $17.64$ & $14.81$	 & $\textbf{83.35}$ & $16.17$ \\
        Longformer Soft Attention Top-K & $89.29$    & $14.23 \pm 2.25$ & $15.57$ & $16.61$	 & $12.46$ & $14.67$ \\
        \hline
        Longformer Ranked Soft Attention & $89.10$    & $22.23 \pm 2.44$ & $\textbf{23.40}$ & $24.51$	 &  $21.33$ & $18.82$ \\
        Compositional Soft Attention & $81.32$ & $21.08 \pm 2.02$ &  $23.19$ & $\textbf{25.21}$	 & $18.78$ & $\textbf{19.67}$ \\
        \hline
    \end{tabular}
    \caption{Full results for FCE GED dataset. Similarly to BEA 2019, we note that the Weighted Soft Attention performs best on the token-level, as evaluated by $F_1$ score. However, that is mainly because the model assigns $1$ to most tokens, causing recall to be high. This is evident by the significantly lower $F_{0.5}$ metric. Our Ranked Soft Attention and Compositional Soft Attention models achieve similar performance, significantly better than Weighted Soft Attention if evaluated on the $F_{0.5}$ and \textit{MAP} metrics.}
    \label{tab:full_fce}
\end{table*}

\begin{table*}[]
    \centering
    \begin{tabular}{l|cccccc}
         & Doc $F_1$ & $F_1$ & $F_{0.5}$ & $P$ & $R$ & MAP \\
         \hline 
        Random Uniform & - & $5.46 \pm 0.18$   & $4.27$ & $3.72$	 &  $10.24$ & $8.46$ \\
        \hline
        Longformer Weighted Soft Attention & $93.41$ &  $6.89 \pm 0.19$ & $4.43$ & $3.58$	 &  $\textbf{92.39}$ &  $7.81$  \\
        Longformer Soft Attention Top-K & $\textbf{94.42}$   & $5.90 \pm 2.21$ &$ 5.19 $& $4.80$	 &  $7.67$ & $7.85$  \\
        \hline
        Longformer Ranked Soft Attention & $92.63$   & $14.13 \pm 2.27$ & $11.92$ & $10.80$	 &  $20.54$ &  $11.97 $\\
        Compositional Soft Attention & $91.85$    & $\textbf{25.46} \pm 0.97$ & $\textbf{20.12}$ & $\textbf{17.66}$	 & $45.85$ & $\textbf{26.82}$ \\
        \hline
    \end{tabular}
    \caption{Full results for the IMDB-Pos Sentiment Detection dataset. We note that our Compositional Soft Attention architecture performs significantly better across all token-level metrics.}
    \label{tab:full_imdb_pos}
\end{table*}

\begin{table*}[]
    \centering
    \begin{tabular}{l|cccccc}
         & Doc $F_1$ & $F_1$ & $F_{0.5}$ & $P$ & $R$ & MAP \\
         \hline 
        Random Uniform & - & $6.02 \pm 0.23$   & $4.80$ & $4.23$	 & $10.46 $& $9.90$ \\
        \hline
        Longformer Weighted Soft Attention & $93.52$  & $8.34 \pm 0.98$ & $5.40$ & $4.37$	 &  $\textbf{93.38}$ &   $10.59$ \\
        Longformer Soft Attention Top-K & $\textbf{94.60}$ & $5.71 \pm 3.63$  &  $5.02 $  & $4.65$	 & $7.40$ & $9.03 $ \\
        \hline
        Longformer Ranked Soft Attention &$ 93.92$  & $19.62 \pm 0.61$ & $ 16.98$  & $15.63$	 &  $27.27$ & $16.44$  \\
        Compositional Soft Attention & $90.82$ & $\textbf{27.27} \pm 3.68$  &  $\textbf{22.70}$ & $\textbf{20.42}$	 & $41.22$ & $\textbf{29.78}$  \\
        \hline
    \end{tabular}
    \caption{Full results for the IMDB-Neg Sentiment Detection dataset. Similarly to IMDB-Pos, we note that our Compositional Soft Attention architecture performs significantly better across all token-level metrics.}
    \label{tab:full_imdb_neg}
\end{table*}

\end{document}